\documentclass[letterpaper, 10 pt, conference]{ieeeconf}
\IEEEoverridecommandlockouts  
\usepackage[utf8]{inputenc}
\usepackage{amsfonts}
\usepackage{amsmath}
\usepackage{hyperref}
\usepackage{tikz}
\usepackage{verbatim}
\usepackage{footnote}
\usetikzlibrary{automata,positioning,shapes,arrows}
\usetikzlibrary{shapes,snakes}
\tikzstyle{block} = [draw, rectangle, minimum size=2em]

\newtheorem{problem}{Problem}
\newtheorem{remark}{Remark}
\newtheorem{example}{Example}
\usepackage{xargs}                      
\usepackage[colorinlistoftodos,prependcaption]{todonotes}
\newcommandx{\answer}[2][1=]{\todo[linecolor=green,backgroundcolor=green,bordercolor=green,#1]{#2}}

\usepackage{environ}
\NewEnviron{myequation}{%
\begin{equation}
\scalebox{1}{$\BODY$}
\end{equation}
}

\begin{document}
\title{\LARGE\bf Reward-Based Deception with Cognitive Bias}

\author{Bo~Wu, Murat Cubuktepe, Suda Bharadwaj, and Ufuk Topcu
	\thanks{ Bo Wu, Murat Cubuktepe, Suda Bharadwaj and Ufuk Topcu are with the Department of Aerospace Engineering
and Engineering Mechanics, and the Oden Institute for Computational
Engineering and Sciences, University of Texas, Austin, 201 E 24th
St, Austin, TX 78712. email: {\tt\small $\{$bwu3, mcubuktepe, suda.b, utopcu$\}$@utexas.edu}}}%

\maketitle
\begin{abstract}
Deception plays a key role in adversarial or strategic interactions for the purpose of self-defence and survival. This paper introduces a general framework and solution to address deception. Most existing approaches for deception consider obfuscating crucial information to rational adversaries with abundant memory and computation resources. In this paper, we consider deceiving adversaries with bounded rationality and in terms of expected rewards. This problem is commonly encountered in many applications especially involving human adversaries. Leveraging the cognitive bias of humans in reward evaluation under stochastic outcomes, we introduce a framework to optimally assign resources of a limited quantity to optimally defend against human adversaries. Modeling such cognitive biases follows the so-called prospect theory from  behavioral psychology literature. Then we formulate the resource allocation problem  as a signomial program to minimize the defender's cost in an environment modeled as a Markov decision process. We use police patrol hour assignment as an illustrative example and provide detailed simulation results based on real-world data.
\end{abstract}

\section{Introduction}
Deception refers to a deliberate attempt to mislead or confuse adversaries so that they may take  strategies that are in the defender's favor \cite{almeshekah2016cyber}. Deception can limit the effectiveness of an adversary's attack, waste adversary's resources and prevent the leakage of  critical information \cite{pawlick2017game}. It is a widely observed behavior in nature for self-defence and survival. Deception also plays a key role in many aspects of human society, such as economics \cite{bonetti1998experimental}, warfare \cite{holt2010deceivers}, game \cite{morgulev2014deception}, cyber security \cite{pawlick2017game} and so on.


In this paper, we focus on the scenario in which the adversary acts in an environment where this interaction is modeled as a Markov decision process (MDP) \cite{puterman2014markov}. The adversary's aim is to collect  rewards at each state of the MDP and the defender tries to minimize the accumulated reward through \emph{deception}. Many existing approaches for deception rely on a rational adversary with sufficient memory and computation power to find its optimal policy \cite{horak2017manipulating,almeshekah2016cyber}. However, deceiving an adversary with only  bounded rationality \cite{simon1957models}, i.e., one whose decisions may follow certain rules that deviate from the optimal action \cite{camerer2011behavioral},  has not been adequately studied so far. Deceiving an adversary with bounded rationality finds, for example, its application in  intrusion detection and protection \cite{yang2014adaptive} or public safety \cite{zhang2014defending}.  Different from obfuscating  sensitive system information to the adversary \cite{8392381,masters2017deceptive}, by deception, we mean that the defender optimally assigns a limited resource to each state, such that the expected cost from defender's perspective (or equivalently, the reward for the adversary) incurred by an adversary can be minimized, even though the adversary is expecting more based on his cognitively biased view of rewards.

To deceive a human more effectively, it is essential to understand the human's cognitive characteristics and what affects his decisions (particularly with stochastic outcomes). Works in behavior psychology, e.g. \cite{kahneman2011thinking}, suggested that humans' decision-making follows intuition and bounded rationality. Empirical evidence has shown that humans tend to evaluate gains and losses differently in decision-making~\cite{tversky1992advances}.  Humans tend to over-estimate the likelihood of low-probability events and underestimate the likelihood of high-probability events in a nonlinear fashion \cite{kahneman2013prospect,tversky1992advances}.  Risk-sensitive measures, such as those in the so-called  prospect theory \cite{kahneman2013prospect}, capture such biases and are widely used in psychology and economics to characterize human preferences. Furthermore, humans tend to make decisions that are often sub-optimal~\cite{norling2004folk}. It is generally believed that such sub-optimality is the result of intuitive decisions or preferences that happen automatically and quickly without much reflection \cite{norling2004folk,kahneman2011thinking}. Human decisions are subject to stochasticity due to the limited computational capacity and inherent noise \cite{reverdy2014modeling}. Consequently, human decisions are often cognitively biased (have a different reward mechanism), probabilistic (have a stochastic action selection policy) and memoryless (only depends on the current state). These are the very characteristics of human decision-making we expect to account for in reward-based deception.

This paper investigates how one can deceive a human adversary by optimally allocating limited resources to minimize his rewards. We model the environment as an MDP to capture the choices available to a human decision-maker and their probabilistic outcomes. We consider opportunistic human adversaries, i.e., they usually do not have significant planning and only act based on immediately available rewards \cite{zhang2014defending}. We describe the human adversary's policy to select different actions following the prospect theory and  bounded rationality \cite{simon1957models}. We model both the adversary's perceived reward and defender's cost (equivalently, the adversary's reward from the defender's point of view) as functions of the  resources available at each state of the MDP. Additionally, we define a subset of the states in the MDP as sensitive states that the human adversary should be kept from visiting.

We then formulate the optimal resource allocation problem as a signomial program (SP) to minimize the defender's cost. SPs are a special form of  nonlinear programming problems, and they are generally nonconvex.  Solving nonconvex NLPs is NP-hard~\cite{hochbaum2007complexity} in general, and a globally optimal solution of an SP cannot be computed efficiently. SPs generalize geometric programs (GP), which can be transformed into  convex optimization problems and then can be solved efficiently~\cite{boyd2007tutorial}. In this paper, we approximate the proposed SP to a GP. In numerical experiments, we show that this approach obtain locally optimal solutions of the SP efficiently by solving a number of GPs. We demonstrate the approach with a problem on the assignment of police patrol hour against opportunistic criminals \cite{zhang2014defending}.

The problem we study is closely related to the Stackelberg security game (SSG) which consists of an attacker and a defender that interact with each other.  In SSG, the defender acts first with limited  resources and then the attackers play in response  \cite{an2017stackelberg}. SSG is a popular formalism to study security problems against human adversaries.  Early efforts  focused on one-shot games where an adversary can only take one move \cite{jain2010software} without considering human's bounded rationality. Then repeated SSG was considered in wildlife security \cite{yang2014adaptive} and fisheries \cite{haskell2014robust} where the defender and the adversary can have repeated interaction. However, neither of these papers considered how a human perceives probabilities, where the existence of nonlinear probability weighting curves is a well-known result in prospect theory \cite{kahneman2013prospect}. Such phenomenon was taken into account in \cite{yang2013improving} and \cite{kar2015game}. But \cite{yang2013improving} only studied one-shot games and \cite{kar2015game} did not consider the adversaries may move from place to place. 

The rest of this paper is organized as the following. We first provide the necessary preliminaries for stochastic environment modeling, human cognitive biases and decision-making in Section \ref{sec:preliminaries}. Then we formulate the human deception problem in terms of resource allocation in Section \ref{sec:Problem Formulation} and show that it can be transformed into a signomial program in Section \ref{sec:Signomial Programming Formulation}. We propose the computational approach to solve the signomial program in Section \ref{sec:Computational Approach for the Signomial Program}.  Section \ref{sec:Experiment} shows simulations results and discusses their implications. We conclude our paper and discusses possible future directions in Section \ref{sec:Conclusion}.

\section{Preliminaries}
\label{sec:preliminaries}
\subsection{Monomials, Posynomials, and Signomials.}\label{def:posy}
  Let $V=\{x_1,\ldots,x_n\}$ be a finite set of strictly positive real-valued \emph{variables}.
  A \emph{monomial} over $V$ is an expression of the form
  \begin{align*}
      f=c\cdot x_{1}^{a_{1}}\cdots x_{n}^{a_{n}}\ ,
  \end{align*}
  where $c\in \mathbb{R}^+ $ is a positive coefficient, and $a_i\in\mathbb{R}$ are exponents for $1\leq i\leq n$. 
  A \emph{posynomial} over $V$ is a sum of one or more monomials:
  \begin{align}
      g=\sum_{k=1}^K c_k\cdot x_1^{a_{1k}}\cdots x_n^{a_{nk}} \ .\label{eq:signomial}
  \end{align}
  If $c_k$ is allowed to be a negative real number for any $1\leq k\leq K$, then the expression~\eqref{eq:signomial} is a \emph{signomial}. 

This definition of monomials differs from the standard algebraic definition where exponents are positive integers with no restriction on the coefficient sign. A sum of monomials is then called a \emph{polynomial}. 

\subsection{Nonlinear programs.}\label{def:nlp}
 A general nonlinear program (NLP) over 
a set of real-valued variables $V$ is
\begin{align}
	\text{minimize} 		&\quad f\label{eq:nl_obj} \\
	\text{subject to} 		&\notag\\
	&\quad g_i \leq 1, \quad i=1,\ldots,m,	\label{eq:nl_ineq}\\
	&\quad h_j = 1, \quad j=1,\ldots,m,\label{eq:nl_eq}
\end{align}
where $f$, $g_i$, and $h_j$ are arbitrary functions over $V$, and $m$ and $p$ are the 
number of inequality and equality constraints of the program respectively.

\subsection{Signomial programs and geometric programs.} 
A special class of NLPs known as \emph{signomial programs} (SP) is of the form~\eqref{eq:nl_obj}--\eqref{eq:nl_eq} where $f$, $g_i$ and $h_j$ are signomials over $V$, see Def.~\ref{def:posy}. 
A geometric program (GP) is an SP of the form~\eqref{eq:nl_obj}--\eqref{eq:nl_eq} where $f,g_i$ are posynomial functions and $h_j$ are monomial functions over $V$. GPs can be transformed into convex programs~\cite[\S{}2.5]{boyd2007tutorial} and then can be solved efficiently using interior-point methods~\cite{boyd2004convex}. SPs are non-convex programs in general, and therefore there is no efficient algorithm to compute global optimal solutions for SPs . However, we can efficiently obtain local optimal solutions for SPs in our setting, as shown in the following sections.

In this paper, the adversary with bounded rationality moves in an environment modeled as a Markov decision process (MDP) \cite{puterman2014markov}. 

\subsection{Markov Decision Processes.}
	A  (MDP) is a tuple $\mathcal{M}=(S,\nu,A,T,U)$ where
	\begin{itemize}
		\item $S$ is a finite set of states;
		\item $\nu:S\rightarrow [0,1]$ is the initial state distribution;
		\item $A$ is a finite set of actions;
		\item $T(s,a,s'):=P(s'|s,a)$. That is, the probability of transiting from $s$ to $s'$ with action $a$; and
		\item $U(s)\in\mathbb{R}^+$ is the utility function that assigns  resources with a quantity $U(s)$ to state $s$.
	\end{itemize}

At each state $s$, an adversary has a set of actions available to choose. 
Then the nondeterminism of the action selection has to be resolved by a policy $\pi$ executed by the adversary. A (memoryless) policy $\pi:S\times A \rightarrow [0,1]$  of an MDP $\mathcal{M}$ is a function that maps every state action pair $(s,a)$ where $s\in S$ and $a\in A$  with probability $\pi(s,a)$.

By definition, the policy $\pi$ specifies the probability for the next action $a$ to be taken at the current state $s$. A bounded rational adversary is often limited in memory and computation power, therefore we only consider the memoryless policies.

In an MDP, a finite state-action path is $\omega=s_0a_0s_1a_1...$, where $s_i \in S,a_i\in A$ and $ T(s_i,a_i,s_{i+1})>0$. Given a policy $\pi$, it is possible to calculate the probability of such path $P_\pi(\omega)$ as 
\begin{equation}\label{eqn:path}
P_\pi(\omega)=\nu(s_0)\prod_i\pi(s_i,a_i)T(s_i,a_i,s_{i+1}).    
\end{equation}

\section{Reward-Based Deception}\label{sec:Problem Formulation}
We assume that an  adversary with bounded rationality moves around in an environment modeled as an MDP $\mathcal{M}=(S,\nu,A,T,U)$. When the adversary is at a state $s\in S$, from the defender's point of view, the immediate reward for the human adversary (or equivalently, the cost for the defender) is 
$$R(s)=g(U(s))\in\mathbb{R}^+,$$
which is a function of allocated resource $U(s)$.  However, due to the bounded rationality and cognitive biases, the perceived immediate reward $R_h(s)$ at state $s$ by the adversary is a different function of $U(s)$, and is given by 
$$
R_h(s)=f(U(s))\in\mathbb{R}^+,
$$
where $f$ is another function over $U$. For a given policy $\pi$, expected rewards $Q_t(s)$ at each state $s$ and time t with a finite time horizon $H$ can be evaluated as
\begin{equation}\label{eqn:reward}
Q_t(s) = R(s)+\sum_{a}\sum_{s'}\pi(s,a) T(s,a,s')Q_{t+1}(s'),
\end{equation}
where $t=0...,H-1$, $Q_{H}(s)=R(s)$. Therefore, $Q_t$ represents the expect accumulated cost of the defender, or equivalently, expected rewards for the human adversary obtained from the policy $\pi$.

The defender's objective is to optimally assign the resources to each state to minimize  his cost (equivalently, the adversary's reward) $Q$, where
\begin{equation}\label{eqn:objective}
Q=\sum \nu(s)Q_0(s),
\end{equation}
by designing the utility function $U$, where the resources are of limited quantity, i.e., $\sum_s U(s)=D$.  Also imagine that there are set of sensitive states $S_s\subset S$ that the adversaries should be kept away from.  Denote  the set of paths that reach $S_s$ in $H$ steps as $\Omega$ such that for each $\omega\in\Omega$ where $\omega=s_0a_0,...,s_N$, we require $N\leq  H, s_i\notin S_s,i<N$ and $s_N\in S_s$. In particular, given a policy $\pi$, $P(\diamondsuit^{\leq H} S_s)$ can be calculated as 
\begin{equation}\label{eqn:reach}
    P(\diamondsuit^{\leq H} S_s)=\sum_{\omega\in\Omega}P(\omega).
\end{equation}

\begin{problem}\label{problem:1}
Given an MDP $\mathcal{M}=(S,\nu,A,T,U)$, time horizon $H$, human reward function defined in (\ref{eqn:CPT1}) and the policy $\pi$, design reward function $U$ with a limited total budget $\sum_s U(s)=D$, such that $Q$ as defined in (\ref{eqn:objective}) can be minimized $\hat{s}$ and $P(\diamondsuit^{\leq H} S_s)\leq\lambda$, which requires that the probability to reach $S_s$ in $H$ steps should be no larger than $\lambda\in[0,1]$, i.e.,
\begin{equation}\label{eqn:reach2}
    P(\diamondsuit^{\leq H} S_s)\leq\lambda.
\end{equation}
\end{problem}

\begin{remark}
Problem \ref{problem:1} studies how to optimally assign the reward to trick the adversary into thinking that his policy could obtain more rewards but in fact, the actual expected reward is minimized with a low probability of visiting sensitive states $S_s$.
\end{remark}

\subsection{Human Adversaries with Cognitive Biases}
To solve Problem \ref{problem:1}, it is essential to find the adversary's policy $\pi$. In this paper, we take human as the adversary with bounded rationality who is opportunistic, meaning that he does not have a specific attack goal nor plans strategically, but is flexible about his movement plan and seek opportunities for  attacks \cite{abbasi2015human}. Those attacks may incur rewards to the human adversary and consequently certain costs for the defender. The process of human decision-making typically follows several steps \cite{doya2008modulators}. First, a human recognizes his current situation or state. Second, he will evaluate each available action based on the potential immediate reward it can bring. Third, he will select an action following some rules. Then he will receive a reward and observe a new state. In this section, we will introduce the modeling framework for the second and third step.

For a human with bounded rationality, the value of a reward from an action is a function of the possible outcomes and their associated probabilities. The prospect theory developed by Kahneman and Tversky  \cite{kahneman2013prospect} is a frequently used modeling framework to characterize the reward perceived by a human.  Prospect theory claims that humans tend to over-estimate the low probabilities and underestimate the high probabilities in a nonlinear fashion.  For example, between winning $100$ dollar with $\frac{1}{100}$ probability and nothing else, or $1$ dollar with probability $1$, humans tend to prefer the former, even though both have the same expectation. 

Given  $X$ as the discrete random variable that has a finite set of outcomes $O$, a general form of prospect theory utility $V(X)$ (i.e. the reward anticipated by a human) is the following. 
\begin{equation}\label{eqn:PT}
V(X) = \sum_{x\in O} v(x)w(p(x)),
\end{equation}
where $v(x)\in\mathbb{R}$ denotes the reward perceived by a human from the outcome $x$. The probability $p(x)$ to get the outcome $x$ is weighted by a nonlinear function $w$ that captures the human tendency to over-estimate low probabilities and under-estimate high probabilities.

The expected immediate reward $r_a(s)$ to perform an action $a$ at state $s$ is
\begin{equation}\label{eqn:expected_reward}
r_a(s)=\sum_{s'}R(s') T(s,a,s').
\end{equation}
However, according to prospect theory, from a human's perspective, the perceived expected immediate reward $r_a^h(s)$ is different. Let $X_{s,a}$ be the random variable for the outcome $O_{s,a}$ of executing action $a$ at state $s$. We have $O_{s,a}=\{x_s'|T(s,a,s')>0\}$ where $x_{s'}$ denotes the event that the state transits from $s$ to $s'$ with an action $a$. The distribution of $X_{s,a}$ is defined as follows.
$$
p(x_{s'})=T(s,a,s'),\forall x_{s'}\in O_{s,a}. 
$$
The human perceived reward $v(x_{s'})$ for the outcome $x_{s'}$ depends on  $U(s')$ received from reaching the state $s'$, which is denoted by 
$$
v(x_{s'})=R_h(s')=f(U(s')).
$$ 
As a result,  $r_a^h(s)$ is denoted by 
\begin{equation}\label{eqn:CPT1}
\begin{split}
  r_a^h(s) &= \sum_{x_{s'}\in O_{s,a}} v(x_{s'})w(p(x_{s'}))\\
           &= \sum_{s'}f(U(s'))w(T(s,a,s')). 
\end{split}
\end{equation}
An empirical form  of $w$ is the following \cite{kahneman2013prospect}. 
\begin{equation}\label{eqn:weighting function}
w(p)=\frac{p^{\gamma}}{(p^\gamma+(1-p)^\gamma)^{\frac{1}{\gamma}}}, \gamma>0.
\end{equation}

	\begin{figure}
		\centering	
\begin{tikzpicture}[shorten >=1pt,node distance=3cm,on grid,auto, bend angle=20, thick,scale=0.7, every node/.style={transform shape}] 
				\node[state] (s0)   {$s_0$}; 
				\node[state] (s1) [left =of s0] {$s_1$}; 
				\node[state] (s2) [above left  = of s0]  {$s_2$}; 
				\node[state] (s3) [above right= of s0] {$s_3$}; 
				\node[state] (s4) [right= of s0] {$s_4$}; 
                \node[text width=3cm] at (-3.5,0) {$U(s_1)=5$};		
                \node[text width=3cm] at (-1.6,2.8) {$U(s_2)=2$};
                \node[text width=3cm] at (2.8,2.8) {$U(s_3)=2.5$};
                \node[text width=3cm] at (5,0) {$U(s_4)=2.5$};
                \draw [->] (0,-1) -- (s0);
				\path[->]
				(s0) edge node {$a,0.1$} (s1) 
				(s0) edge node {$a,0.9$} (s2) 
				(s0) edge node {$b,0.5$} (s3) 
				(s0) edge node {$b,0.5$} (s4) 
				; 

				\end{tikzpicture} 
		\caption{A simple example for sub-optimality with human cognitive biases}\label{fig:example}
	\end{figure}
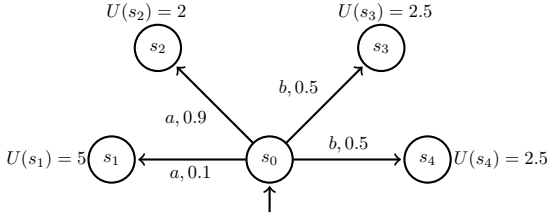
	
	Given an MDP as depicted in Figure \ref{fig:example}, where $S=\{s_0,\ldots,s_4\}$, $A=\{a,b\}$. We assume that $R(s)=U(s)$, $R_h(s)=U(s)^{0.88}$, $\gamma=0.6$ in (\ref{eqn:weighting function}). 
	It can be found from (\ref{eqn:expected_reward}) and (\ref{eqn:CPT1})  that $r_a(s_0)=2.3, r_a^h(s_0)=2.0678, r_b(s_0)=2.5$ and $r_b^h(s_0)=1.8617$. Since $r_a^h(s_0)>r_b^h(s_0)$, suppose a human is at $s_0$, from human's perspective, he will prefer the action $a$.  However, $r_a(s_0)<r_b(s_0)$ which indicates that action $a$  actually has more expected immediate rewards. 

\begin{remark}
In this example, the rewards are already given, and it can be seen that the human could make a sub-optimal decision. It illustrates how  cognitive bias can deviate the human behavior from optimal. 
\end{remark}

After evaluating the outcome of each candidate action $a$ by $r_a^h(s)$, a human then needs to make an action selection. Humans are known to only have quite limited cognitive capabilities. Human's policy $\pi$ to choose an action can be described as a random process that biases toward the actions of high   $r_a^h(s)$, such that
\begin{equation}\label{eqn:mu}
\pi(s,a)=\frac{r_a^h(s)}{\sum_{a'}r_{a'}^h(s)},
\end{equation}
where $\pi(s,a)$ denotes the probability of executing the action $a$ at state $s$. Such a bounded rational behavior has been observed in humans, such as urban criminal activities \cite{short2008statistical}.  Intuitively, it implies that human selects the action $a$ opportunistically at each state $s$ with the probability proportional to the perceived immediate reward $r^h_a(s)$. 

Now we are ready to redefine Problem \ref{problem:1} as follows. 
\begin{problem}\label{problem:2}
Solve Problem \ref{problem:1} for $\pi$ defined as (\ref{eqn:mu}).
\end{problem}

\section{Signomial Programming Formulation}\label{sec:Signomial Programming Formulation}

Given an MDP $\mathcal{M}$, time horizon $H$, human reward function and policy as defined in (\ref{eqn:CPT1}) and (\ref{eqn:mu}), the solution of the Problem 1 can be computed by solving the following signomial program. The $g$ and $f$ are assumed to be monomial functions of $U$ for our solution method. 

	 	\begin{align}
	 	&		\quad	\text{minimize} \quad  Q=\sum \nu(s)Q_0(s) \label{eq:objective}\\
		&	\quad \text{subject to} \nonumber &
\\
	&	\quad \displaystyle		\forall s\in S, t \in  \{0,\ldots,H-1\}, \nonumber \\ &\quad Q_t(s) \geq R(s)+ \sum_{a \in A}\sum_{s' \in S}\pi(s,a) T(s,a,s')Q_{t+1}(s')\label{eq:risk_neutral}\\
		&	\quad \displaystyle		\forall s\in S, t \in  \{0,\ldots,H-1\}, \nonumber\\&\quad P_t(s) \geq \sum_{a \in A}\sum_{s' \in S}\pi(s,a) T(s,a,s')P_{t+1}(s')\label{eq:risk_neutral_probability}\\
		&\quad	\forall t \in  \{0,\ldots,H\}, \forall s\in S_s, \quad P_t(s)=1\label{eq:reachstate}\\
				&\quad \displaystyle	\forall t \in  \{0,\ldots,H\},  \quad \sum_{s\in S}\nu(s)P_t(s)\leq \lambda\label{eq:reachcons}\\
	& \quad	\forall s\in S, a\in A, \nonumber\\&\quad  \pi(s,a) \sum_{a' \in A}\sum_{s'\in S} f(U(s'))w(T(s,a',s')) \nonumber\\ & =  \sum_{s' \in S} f(U(s'))w(T(s,a,s'))\label{eq:policy_computation}\\
		& \quad	\forall s\in S, \quad  R(s)=g(U(s))\label{eq:utility_computation}\\
			&\quad	\sum_{s \in S} U(s)=D, \label{eq:reward_cons}
\end{align}
where variables $R(s)$ are for rewards in each state $s$, $U(s)$ are for utilities in each state $s$, $\pi(s,a)$ are for the probability of taking action $a$ in state $s$ are for each state and action, $Q_t(s)$ are for the expected reward of the state $s$ and time step $t$, and $P_t(s)$ are for the probability of reaching the set of target states $S_s$ in each state $s$ and time step $t$.

The objective in~\eqref{eq:objective} minimizes the accumulated expected reward from the initial state distribution $\nu(s)$ over a time horizon $H$. In~\eqref{eq:risk_neutral}, we compute $Q_t(s)$ by adding the immediate reward in state $s$ and the expected reward of the successor states according to the policy variables $\pi(s,a)$ for each action $a$. The probability of reaching each successor state $s'$ depends on the policy variables $\pi(s,a)$ in each state $s$ and action $a$. Similar to the constraint in~\eqref{eq:risk_neutral}, the variables $P_t(s)$ are assigned to the probability of reaching the set of target states $S_s$ from state $s$ and time step $t$ in~\eqref{eq:risk_neutral_probability}.


The probability of reaching any state $s\in S_s$ in each horizon from the states in $S_s$ is set to 1 as in~\eqref{eq:reachstate}. The constraint in~\eqref{eq:reachcons} assures that the probability of reaching any state $s \in S_s$ from the initial state distribution $\nu(s)$ is less than $\lambda$. The constraint in~\eqref{eq:policy_computation} computes the policy using the model in~\eqref{eqn:mu}. We give the relationship between rewards and utilities in~\eqref{eq:utility_computation}. Finally,~\eqref{eq:reward_cons} gives the total budget for utilities.


The constraint in~\eqref{eq:risk_neutral} and~\eqref{eq:risk_neutral_probability} are convex constraints, because the functions in the right hand sides are posynomial functions, and the functions in the left hand sides are monomial functions. The constraints in~\eqref{eq:reachstate} and~\eqref{eq:reachcons} are affine constraints, therefore they are convex. The constraints in~\eqref{eq:policy_computation} and~\eqref{eq:reward_cons} are equality constraints with posynomials, therefore they belong to the class of signomial constraints, and they are not convex. In the literature, there are various methods to deal with the nonconvex constraints to obtain a locally optimal solution including sequential convex programming, convex-concave programming, branch and bound or cutting plane methods~\cite{moore1991global,boyd2007tutorial,lawler1966branch}.

\section{Computational Approach for the Signomial Program}\label{sec:Computational Approach for the Signomial Program}

In this section, we discuss how to compute a locally optimal solution efficiently for Problem~\ref{problem:1} by solving the signomial program in~\eqref{eq:objective}--\eqref{eq:reward_cons}.
We propose a \emph{sequential convex programming} method to compute a local optimum of the signomial program in~\eqref{eq:objective}--\eqref{eq:reward_cons}, following~\cite[\S{}9.1]{boyd2007tutorial}, solving a sequence of GPs. We obtain each GP by replacing signomial constraints in equality constraints of the SGP~signomial program in~\eqref{eq:objective}--\eqref{eq:reward_cons} with \emph{monomial approximations} of the functions.

\subsection{Monomial approximation}\label{def:approx}
Given a posynomial $f$, a set of variables $ \{x_1,\dots,x_n\}$, and an initial point $\hat{x}$, a \emph{monomial approximation} \cite{boyd2007tutorial} $\hat{f}$ for $f$ around $\hat{x}$ is 
\begin{align*}
	\forall i. 1 \leq i \leq n
	\quad
	\hat f = f[\hat{x}]
	\prod_{i=1}^{n}\Bigg(\dfrac{x_{i}}{\hat{x}(x_i)}\Bigg)^{a_{i}},\\
	\quad
	\text{where }
	a_{i}=\dfrac{\hat{x}(x_i)}{f[\hat{x}]}\dfrac{\partial f}{\partial x_{i}}[\hat{x}].
\end{align*}	

Intuitively, a monomial approximation of a posynomial $f$ around an initial point $\hat{x}$ corresponds to an affine approximation of the posynomial $f$. Such an approximation is provided by the first
order Taylor approximation of $f$, see~\cite[\S{}9.1]{boyd2007tutorial} for more details.


%
%
%
%
%
%
%
%
For a given instantiation of the utility and policy variables $U(s)$ and $\pi(s,a)$, we approximate the SP in~\eqref{eq:objective}--\eqref{eq:reward_cons} to obtain a GP as follows.
We first normalize the utility values to ensure that they sum up to $D$. Then, using those utility values, we compute the policy according to constraint in~\eqref{eq:policy_computation}. After the policy comptutation, we compute a monomial approximation of each posynomial term in the constraints \eqref{eq:policy_computation} and \eqref{eq:reward_cons} around the previous instantiation of the utility and policy variables. After the approximation, we solve the approximate GP. We repeat this procedure until the procedure converges. 


One key problem with this approach is, we require an initial feasible point to the signomial problem in~\eqref{eq:objective}--\eqref{eq:reward_cons}, which may be hard to find because of the reachability constraint in~\eqref{eq:reachcons}. Therefore, we introduce a new variable $\tau$ and we replace the reachability constraint in~\eqref{eq:reachcons} by the following constraints:
	 	\begin{align}
				&\displaystyle\quad	\forall t \in  \{0,\ldots,H\},  \quad \sum_{s \in S}\nu(s)P_t(s)\leq \lambda\cdot \tau\label{eq:reachconsrelax}\\
		&\quad \tau \geq 1. \label{eq:reachconsrelax2}
\end{align}
By replacing the reachability constraint, we ensure that any initial utility function and policy is feasible to the signomial program in~\eqref{eq:objective}--\eqref{eq:reachconsrelax2}. To enforce the feasability of the reachability constraint in~\eqref{eq:reachcons}, we change the objective in~\eqref{eq:objective} as follows:
	 	\begin{align}
	 	&		\quad	\text{minimize} \quad  Q+\delta\cdot\tau \label{eq:objective2}
\end{align}
where $\delta$ is a positive \emph{penalty parameter} that determines the violation rate for the \emph{soft} constraint in~\eqref{eq:reachconsrelax}. In our formulation, we increase $\delta$ after each iteration to satisfy the reachability constraint.

%



We stop the iterations when the change in the value of $Q$ is less than a small positive constant $\epsilon$. Intuitively, $\epsilon$ defines the required improvement on the objective value for each iteration; once there is not enough improvement, the process terminates.

\section{Numerical Experiment}\label{sec:Experiment}

Let us consider an urban security problem, where a criminal plans his next move randomly based on his local information on the nearby locations that are protected by police patrols. Such a criminal is opportunistic, i.e, he is not highly strategic by conducting careful surveillance and rational planning before making moves. It is known that this kind of opportunistic adversaries contribute to the majority of the urban crimes \cite{brantingham2008offender}.   For prevention and protection, each location should be assigned a certain police patrol hours. Due to the limited amount of police resources, the total number of patrol hours is  limited as well.

\begin{figure}[h]
    \centering
    \includegraphics[scale=0.49]{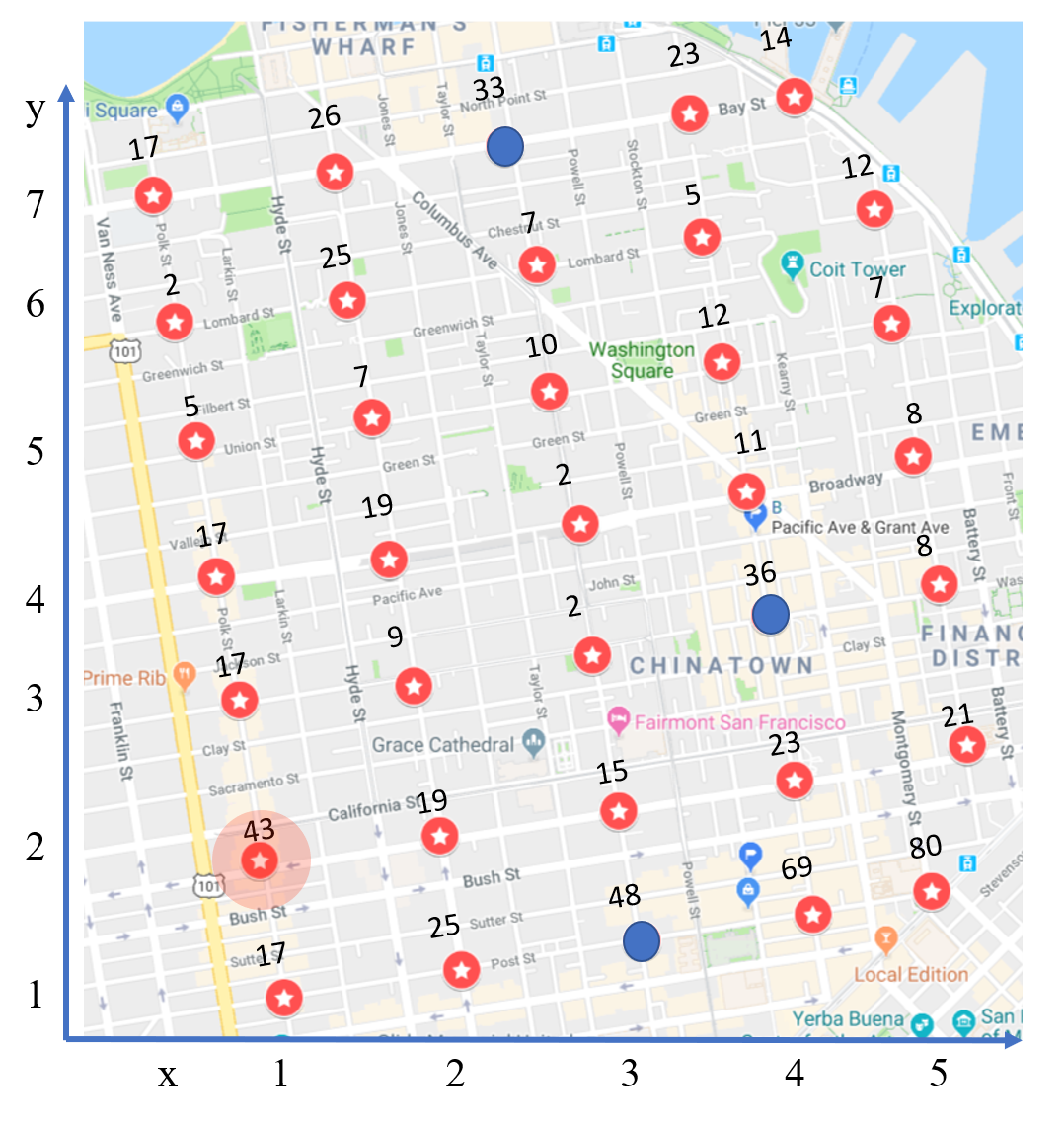}
    \caption{The $35$ intersections in North East of San Francisco. The  map is obtained from Google map.  The shaded area is a circle with a 500 feet radius. The numbering of the states starts from the bottom left corner and goes from left to right in every row. The number beside each location indicates the number of crimes in that area. }
    \label{fig:crime map}
\end{figure}

Figure \ref{fig:crime map} shows $35$ intersections in San Francisco, CA with $7$ rows and $5$ columns.  We use an MDP $\mathcal{M}=(S,\nu,A,T,U)$ to describe the network of the set of intersections $S$. The number $C(s)$ of crimes  that occurred in the first four weeks of October, 2018 within 500 feet of each interaction $s$ is shown in Figure \ref{fig:crime map}. The crime data are obtained from https://www.crimemapping.com/map/ca/sanfrancisco. The criminal can choose to move left, right, up or down to the immediate neighboring intersections. Consequently, there are four actions available. The execution of each action will lead the human to its intended neighborhood of the intersection with a high probability ($\geq 0.95$) and small probability to other neighboring intersections to account for unexpected change of movement plan.  

Initially, the criminal has equal probability to appear at any state, i.e., $\nu(s)=\frac{1}{35}$ for any $s\in S$. The utility $U(s)$ denotes the number of police patrol hours that should be allocated to the vicinity of each intersection. The total number of police patrol hours is $D=\sum U(s)=400$.   If a location $s$ is assigned with $U(s)$ patrol hours, its reward to the criminal  (equivalently, the cost to the defender) is $$R(s)=\frac{C(s)}{U(s)}.$$ 
Intuitively, it means that the reward to the human adversary, from the defender's point of view, is proportional to the crime rate indicated by $C(s)$ and inversely proportional to the police patrol hours. The reward from the human adversary's view is evaluated as
$$f(U(s))=R(s)^{0.88},$$
which is a function commonly seen in the literature to describe how human biases the reward \cite{tversky1992advances}.

Initially, the criminal is at  $s$ with probability $\nu(s)$, where he tries to plan his move over the next $H$ steps. The objective is to  assign the police patrol hours to each state, such that the expected accumulated reward in $H$ steps received by the criminal is minimized. The sensitive states $S_s=\{3,14,33\}$ should be visited with a probability no larger than $\lambda=0.3$, i.e.
$$
P(\diamondsuit^{\leq H} S_s)\leq 0.3.
$$
The sensitive states are also shown as blue circles in Figure \ref{fig:crime map}.
\begin{figure}[t]
\centering

\includegraphics[scale=0.42]{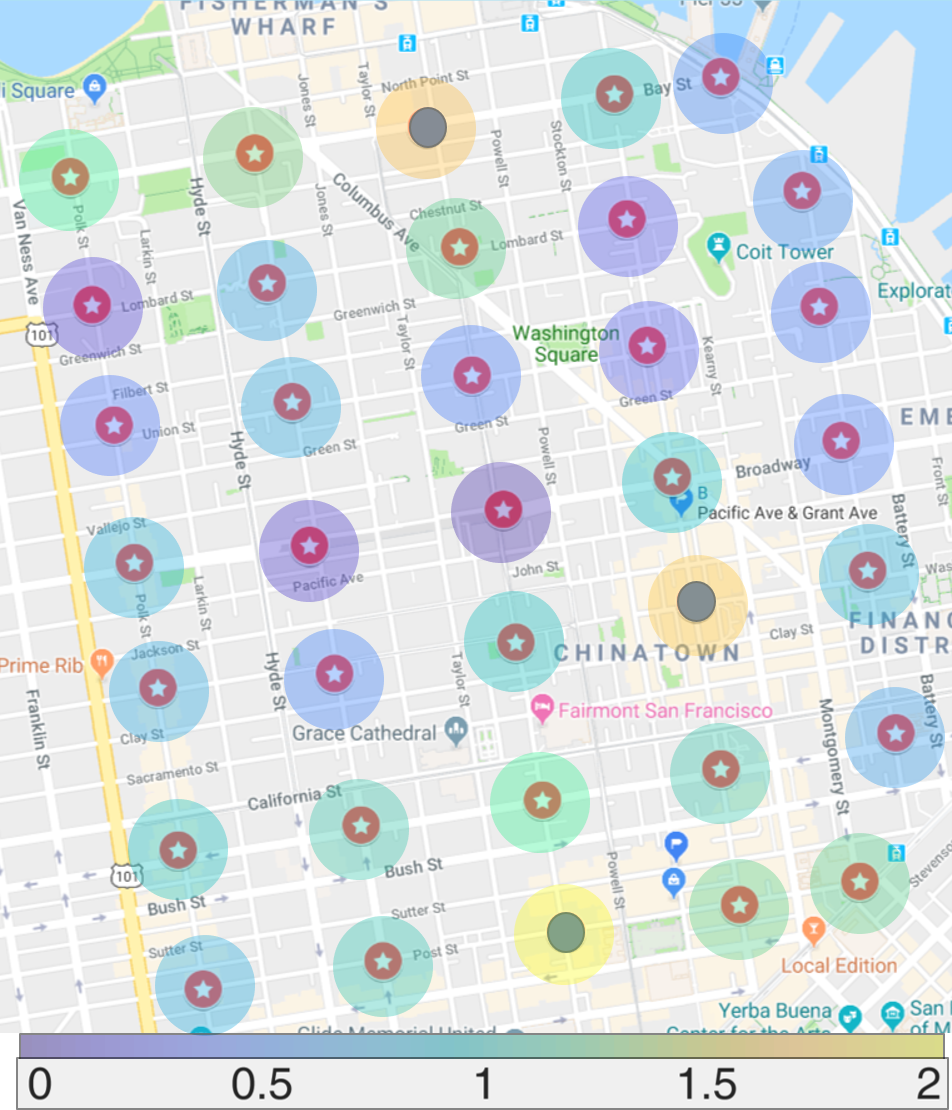}

		\caption{The distribution of utility $U$.} \label{fig:U}
\end{figure}

We formulate the problem as a signomial program. From an initial uniform utility distribution, we instantiate the policies and reward functions. Then, from the initial values, we linearize the signomial program in~\eqref{eq:objective}--\eqref{eq:objective2} to a geometric program. We parse the geometric programs using the tool GPkit~\cite{gpkit}, and solve them using the solver MOSEK~\cite{mosek}. We set $\epsilon=10^{-4}$ for convergence tolerance. All experiments were run on a 2.3 GHz machine with 16 GB RAM. The procedure converged after 32 iterations for a problem with horizon length $T=20$ in 230.06 seconds. The expected reward $Q$ from the initial state distribution is 117.15, and the reachability probability of the sensitive states from the initial state distribution is $0.192$, which satisfies the reachability specification.

The result is  shown in Figure \ref{fig:U}. Different colors at each intersection show the number of patrol hours, i.e, the resource $U(s)$, assigned to each location s. In Figure \ref{fig:U}, $U(s)$ is shown with a logarithmic scale for better illustration. As the color bar at the bottom of the figure indicates, the closer the color at each location is to the right side of this bar, the higher patrol hours are assigned. For example, the state at $(3,1)$ (the third state from the first row), where $C(s)=48$ gets assigned patrol hour equals $106$ which is approximately $2$ in logarithmic scale. Therefore, its color is yellow in Figure \ref{fig:U} as indicated by the right tip of the color bar. Together with Figure  \ref{fig:crime map}, it can be observed that sensitive places and places with a higher number of crimes get assigned more patrol hours.  Consequently, the rewards at those states are fairly low  to discourage the criminal from visiting it. The cost at each location is proportional to the crime rate and inversely proportional to the police patrol hours. The patrol hours assigned to each place intends to minimize the expected cost incurred by the human adversary.

\section{Conclusion}\label{sec:Conclusion}
This paper introduces a general framework for deceiving adversaries with bounded rationality in terms of the obtained reward minimization. Leveraging the cognitive bias of the human from well-known prospect theory, we formulate the reward-based deception  as a resource allocation problem  in Markov decision process environment and solve as a signomial program to minimize the adversary's expected reward. We use police patrol hour assignment as the illustrative example and show the validity of our propose solution approach. It opens doors for further research on the topic to consider the scenarios where defender can move around and react to the human adversaries in real time, and the human adversary has a learning capability to adapt the defender's deceiving policy.

\bibliographystyle{IEEEtran}
\bibliography{ref}
\end{document}